\pdfoutput=1

\documentclass{article} 
\usepackage{amsmath}
\usepackage{booktabs}
\usepackage[dvipsnames]{xcolor}
\usepackage{graphicx}
\usepackage[pdftitle={E-swish: Ajusting Activations to Different Network Depths}]{hyperref}
\usepackage{times}
\usepackage{url}
\usepackage{subcaption} 

\usepackage[backend=biber]{biblatex}

\graphicspath{}

\title{E-swish: Adjusting Activations to Different Network Depths}

\author{Eric Alcaide\\
Independent Researcher, Barcelona \\
\texttt{ericalcaide1@gmail.com} \\
}
\date{January, 2018}

\begin{document}

\maketitle

\begin{abstract}
\noindent
Activation functions have a notorious impact on neural networks on both training and testing the models against the desired problem. Currently, the most used activation function is the Rectified Linear Unit (ReLU). This paper introduces a new and novel activation function, closely related with the new activation \(Swish = x*sigmoid(x)\) (Ramachandran et al., 2017) \cite{ramachandran} which generalizes it. We call the new activation \(E-swish = \beta x*sigmoid(x)\).
\\
We show that E-swish outperforms many other well-known activations including both ReLU and Swish. For example, using E-swish provided 1.5\% and 4.6\% accuracy improvements on Cifar10 and Cifar100 respectively for the WRN 10-2 when compared to ReLU and 0.35\% and 0.6\% respectively when compared to Swish.
The code to reproduce all our experiments can be found at \url{https://github.com/ EricAlcaide/E-swish} 
\end{abstract}

\section{Introduction}
\label{sec:introduction}

The election of the activation function has a notorious impact on both training and testing dynamics of a Neural Network. A correct choice of activation function can speed up the learning phase and it can also lead to a better convergence, resulting in an improvement on metrics and/or benchmarks.
\\
Initially, since the first neural networks where shallow, sigmoid or tanh nonlinearities were used as activations. The problem came when the depth of the networks started to increase and it became more difficult to train deeper networks with these functions (Glorot and Bengio, 2010)\cite{glorot_and_bengio}.
\\
The introduction of the Rectifier Linear Unit (ReLU) (Hahnloser et al., 2000\cite{hahnloser_et_al}; Jarrett et al., 2009\cite{jarrett_et_al}; Nair \& Hinton, 2010\cite{nair_and_hinton}), allowed the training of deeper networks while providing improvements which allowed the accomplishment of new State-of-the-Art results (Krizhevsky et al., 2012\cite{krizhevsky_et_al}).
\\
Since then, a variety of activations have been proposed (Maas et al., 2013\cite{maas_et_al}; Clevert et al., 2015\cite{clevert_et_al}; He et al., 2015\cite{he_linear}; Klambauer et al., 2017\cite{klambauer_et_al}). However, none of them have managed to replace Relu as the default activation for the majority of models due to inconstant gains and computational complexity (Relu is simply max(0, x)).
\\
In this paper, we introduce a new activation function closely related to the recently proposed Swish function (Ramachandran et al., 2017\cite{ramachandran}), which we call E-swish. E-swish is just a generalization of the Swish activation function (x*sigmoid(x)), which is multiplied by a parameter β:
\\
\begin{equation} \label{e_swish}
f\left(x\right)=\beta x\cdot sigmoid(x)
\end{equation}
\\
Our experiments show that E-swish systematically outperforms any other well-known activation function, providing not only a better overall accuracy than both Relu and Swish and sometimes matching the speed of Elu. Our experiments show that E-swish outperforms both Relu and Swish even when the hyperparameters are designed for Relu. For example, in the Wide ResNet 16-4 (Zagoruyko \& Komodakis, 2016\cite{zagoruyko_and_komodakis}), E-swish provided an improvement of 1.5\% relative to relu and 0.5\% relative to swish on the Cifar100 dataset.

\section{E-swish}
We propose a new activation function, which we call E-swish:
\\
\[f\left(x\right)=\beta x\cdot sigmoid(x)\]
\\
Where \(sigmoid \left (x \right )=1 / (1+ \exp (-x) \)
\begin{figure}[h!]
\centering
\includegraphics[scale=0.475]{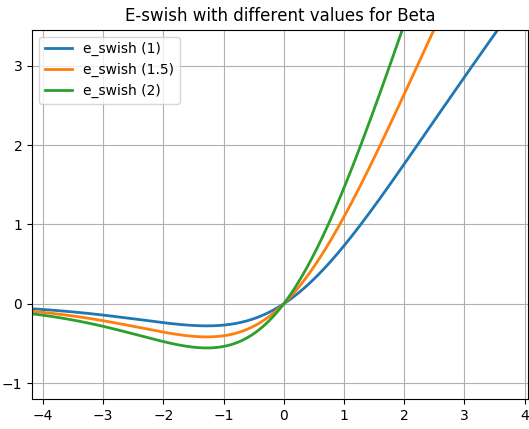}
\caption{E-swish activation function for different values of \(\beta\).}
\label{fig:e_swish}
\end{figure}
The properties of E-swish are very similar to the ones of Swish. In fact, when \(\beta\)=1, E-swish reverts to Swish. Like both Relu and Swish, E-swish is unbounded above and bounded below. Like Swish, it is also smooth and non-monotonic. The property of non-monotonicity is almost exclusive of Swish and E-swish.
Another exclusive feature of both, Swish and E-swish, is that there is a region where the derivative is greater than 1.
\\
As it can be seen in the image above, the maximum slope of the function, and therefore its derivative grows linearly with β. Therefore, we speculate a large coefficient of β my cause gradient exploding problems, while a small one may cause gradient vanishing. Experimentally, we show that a great choice of the parameter β might be \(1 \leq \beta \leq 2\).
\\
\\
The derivative of Swish is: \(f^\prime\left(x\right)=\ f\left(x\right)+\ \sigma\left(x\right)\left(1 - f\left(x\right)\right)\) 
\\
Where \(\sigma\left(x\right)=sigmoid\left(x\right)\ = 1/(1+\exp(-x))\).
\\
\\
The derivative of E-swish, therefore, is:
\\
\begin{equation}
\label{e_swish_deriv}
\begin{split}
f^\prime\left(x\right) & =\ \beta\sigma\left(x\right)+\beta\ x\ \cdot \sigma (x)(1 - \sigma x)=   
\\
& =\ \beta\sigma\left(x\right)+\ \beta x\ \cdot \sigma x- \beta x \cdot \sigma (x)^2 =      
\\
& =\beta x\ \cdot \sigma x + \sigma x\beta - \beta x \cdot \sigma x=  
\\
& =f\left(x\right)+\sigma\left(x\right)\left(\beta-f\left(x\right)\right)
\end{split}
\end{equation}
\\
As can be seen, the two derivatives are very similar, the only change is the parameter β instead of the 1.
\\
As it can be inferred from these formulas, the Figure 1, and as it can be seen in Figure 2, the derivative of E-swish is often bigger than 1 in some regions, especially if the β parameter is high. This may be confusing, but it provides a faster learning and we show experimentally that both E-swish and Swish are able to train deeper networks than Relu when using Batch Normalization (Ioffe \& Szegedy, 2015\cite{ioffe_szegedy}).
\\
\begin{figure}[h!]
\centering
\includegraphics[scale=0.4]{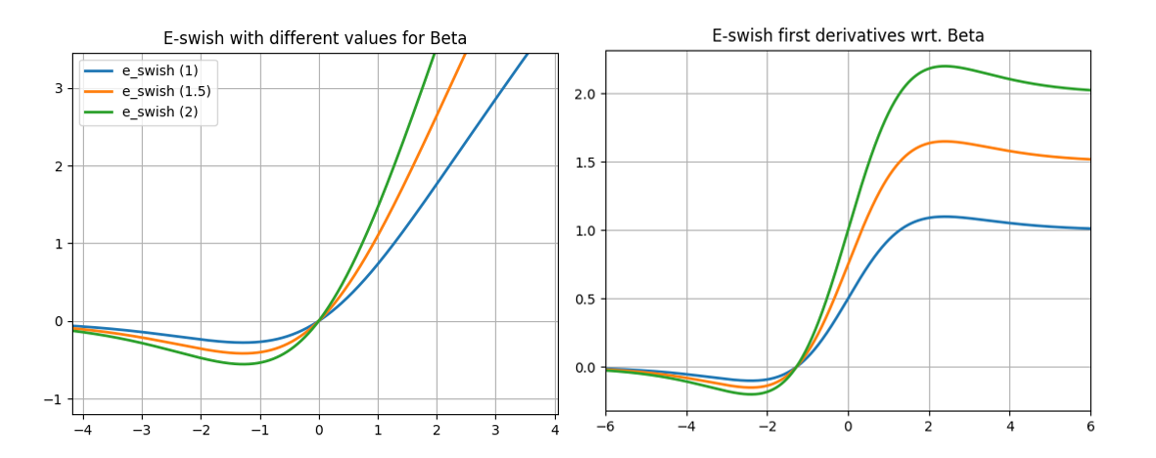}
\caption{First and derivatives of E-swish with respect to \(\beta\).}
\label{fig:e_swish_derivs}
\end{figure}
\\
E-swish can be implemented as a custom activation in some popular deep learning libraries (eg. \(\beta\)*x*K.sigmoid(x) when using Keras or \(\beta\)*tf.nn.swish(x)when using Tensorflow…). Implementations of E-swish in some widely used deep learning frameworks will be provided together with the code to reproduce the experiments performed in this paper.
\\
\subsection{Properties of E-swish}
This paper takes as a reference the work of the original Swish paper (Ramachandran et al., 2017\cite{ramachandran}). For that reason, we compare E-swish to Relu and Swish as we consider them the basis to work on and compare to. We don’t provide extensive comparisons with other activation functions, since they’re provided in the original Swish paper.
\begin{figure}[h!]
\centering
\includegraphics[scale=0.5]{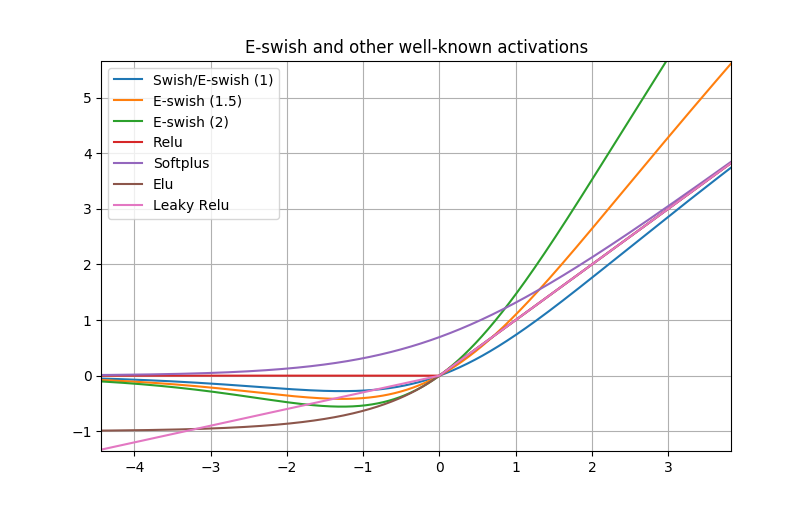}
\caption{E-swish for \(\beta\)=1.5 and 2 plus other well-known activation functions. Best viewed in color.}
\label{fig:activations}
\end{figure}
\\
It’s very difficult to determine why some functions perform better than others given the presence of a lot of confounding factors. Despite of this, we believe that the non-monotonicity of E-swish favours its performance. The fact that the gradients for the negative part of the function approach zero can also be observed in Swish, Relu and Softplus activations. However, we believe that the particular shape of the curve described in the negative part, which gives both Swish and E-swish the non-monotonicity property, improves performance since they can output small negative numbers, unlike Relu and Softplus.
\\
\\
We also believe that multiplying the original Swish function by β>1 accentuates its properties. However, choosing a large β may cause gradient exploding problems. For that reason, we conduct our experiments with values for the parameter β in the range \(1 \leq \beta \leq 2\).
\\
\\
Inspired by the original Swish paper, we plot below the output landscape of a random network, which is the result of passing a grid in form of points coordinates. The network is composed of 6 layers with 128 neurons each and initialized to small values using the Glorot uniform initialization (Glorot \& Bengio, 2010\cite{glorot_and_bengio}). 
\\
\\
\begin{figure}[h!]
\centering
\includegraphics[scale=0.5]{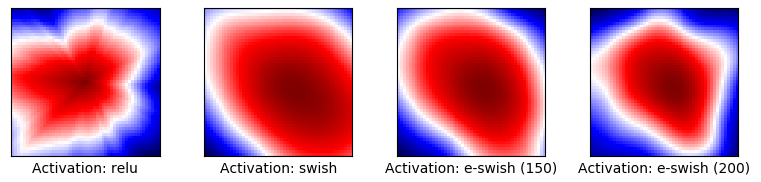}
\caption{Output landscape for a random network with different activation functions. Best viewed in color.}
\label{fig:lanscapes_2d}
\end{figure}
\\
In the original Swish paper, the improvements associated with the Swish function were supposed to be caused by the smoothness of the Swish output landscape, which directly impacts the loss landscape. As can be seen in Figure 4, E-swish output landscapes also have the smoothness property.
\\
If we plot the output landscapes in 3D, as can be seen in Figure 5, it can be observed that the slope of the E-swish landscape is higher than the one in Relu and Swish.  We found that the higher the parameter β, the higher the slope of the loss landscape. Therefore, we can infer that, probably, E-swish will show a faster learning than Relu and Swish as the parameter β increases. As a reference, we also plot the landscape of the Elu activation function (Clevert et al., 2015\cite{clevert_et_al}) since it has been proved to provide a faster learning than Relu, and its slope is higher too. 

\begin{figure}[h!]
\centering
\includegraphics[scale=0.375]{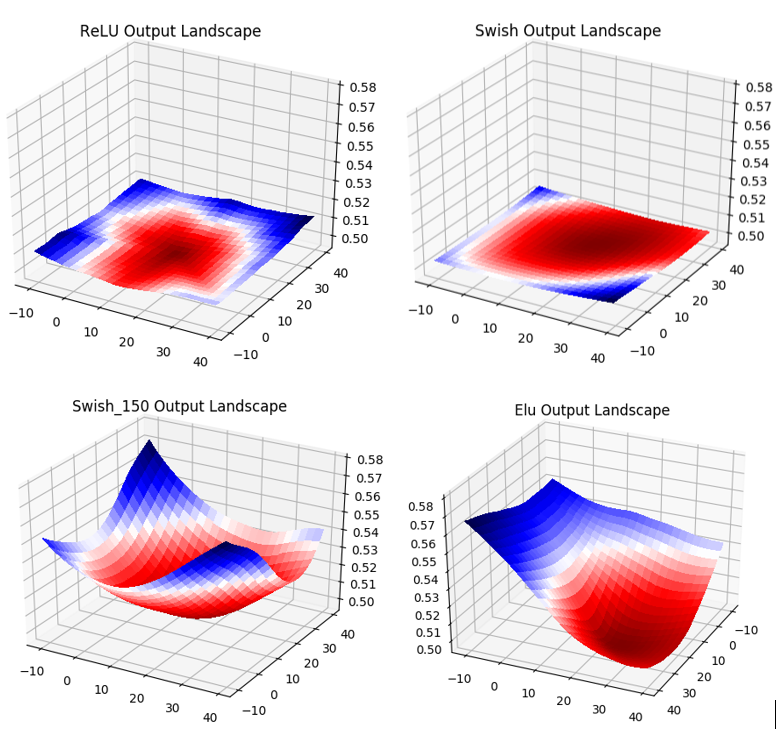}
\caption{3D projection of the output landscape of a random network. Values were obtained by fitting the coordinates of each point in a grid to a 6-layer, with 128 neurons each, randomly initialized neural network. Best viewed in color.}
\label{fig:lanscapes_3d}
\end{figure}
Figures 4 and 5 are very similar to the one presented in Ramachandran et al., 2017\cite{ramachandran}.

\section{Experiments}
\noindent
In this section we will provide a detailed explanation of our tests for the E-swish activation function. To measure the performance of E-swish relative to other activations, we benchmark the results obtained in the following datasets: MNIST, CIFAR10 and CIFAR100. All our experiments were run on a Nvidia GeForce 1060Ti with 6Gb of memory.

\subsection{Training Deep Networks}
First of all, we want to prove that E-swish has also the ability to train deeper networks than Relu, as shown for Swish in Ramachandran et al., 2017\cite{ramachandran}.  We conduct a similiar experiment to the one performed in the paper mentioned before.
\\
\\
We train fully connected networks of different depths on MNIST with 512 neurons each layer. We use SGD as optimizer, the initial learning rate is set to 0.01 and momentum to 0.9. We multiply the learning rate by 0.35 whenever there are 2 epochs with no improvement in the validation accuracy. We train each network for 15 epochs although we terminate the training earlier in case that there is no improvement in the validation accuracy for 5 epochs. We use Glorot uniform initialization (Glorot \& Bengio, 2010\cite{glorot_and_bengio}) and a batch size of 128. No dropout is applied.
\\
\\
We don’t use residual connections, since they would allow to train deeper networks. We use Batch Normalization when the last layer’s index (starting at 0) satisfies that i\%3=1 where i is the index and \% is the modulus operator. Our experiment shows that E-swish performs very similar to Swish in very deep networks when \(\beta \leq 1.5\). We speculate that it goes in line with our hypothesis that large values for \(\beta\) would cause gradient exploding.

\begin{figure}[h!]
\centering
\includegraphics[scale=0.4]{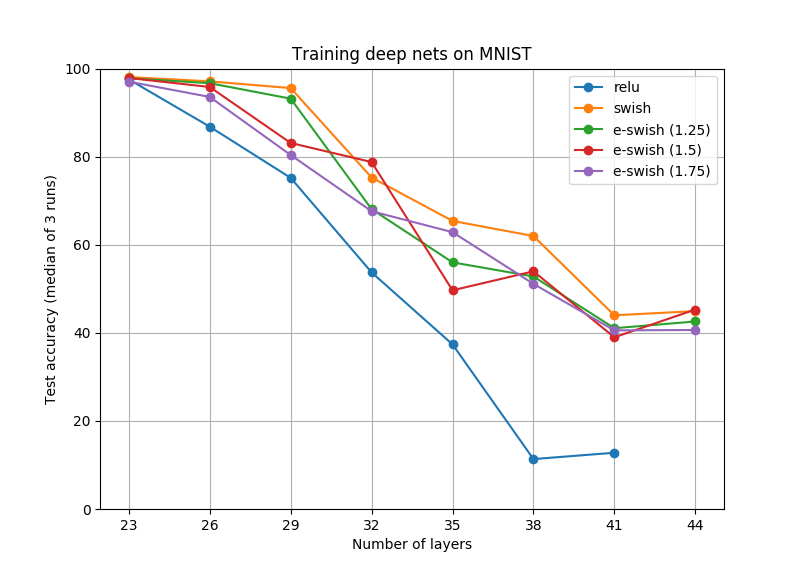}
\caption{Test accuracy (median of 3 runs) on MNIST for fully connected networks with different activations which depths go from 23 to 44 layers. Best viewed in color.}
\label{fig:mnist_single}
\end{figure}
\noindent
\\
In our experiments, the three activations perform almost equally up to 23 layers. Then, we can see a clear divergence between Relu and  Swish / E-swish, where both Swish and E-swish outperform Relu by a large margin in the interval of 23-50 layers. As seen in the Figure 6 above, performance of E-swish is affected when the β parameter is changed.
\\
This experiment proves that both Swish and E-swish outperform Relu when training very deep networks since they achieve better test accuracies.
\\
\subsection{Comparing E-swish to Relu and Swish}
We benchmarked E-swish against the following activation functions:
\begin{itemize}
  \item ReLU (\emph{ReLU}): \( f(x) = max(0, x) \)
  \item Swish (\emph{Swish}): \( f(x) = x \cdot sigmoid(x) \)
\end{itemize}
First, we conduct some small experiments to show the improvements provided by            E-swish on a range of models and architectures.
\\
Our experiments showed that E-swish consistently outperforms both Relu and Swish, while Swish showed inconsistent improvements against Relu. We also noticed that E-swish provides a faster learning than Swish. We conducted our experiments on both single model performance and a median of 3 runs.

\subsubsection{MNIST}
For MNIST, we first compare the different activations on a single model performance. We train a fully connected network of 5 layers with the following number of neurons: 200, 100, 60, 30, 10, with dropout of 0.2. We use SGD with learning rate set to 0.1 and no momentum, we also use Glorot uniform initialization (Glorot \& Bengio, 2010\cite{glorot_and_bengio}) and a batch size of 64. We train each network for 20 epochs. 

\begin{figure}[h!]
\centering
\includegraphics[scale=0.35]{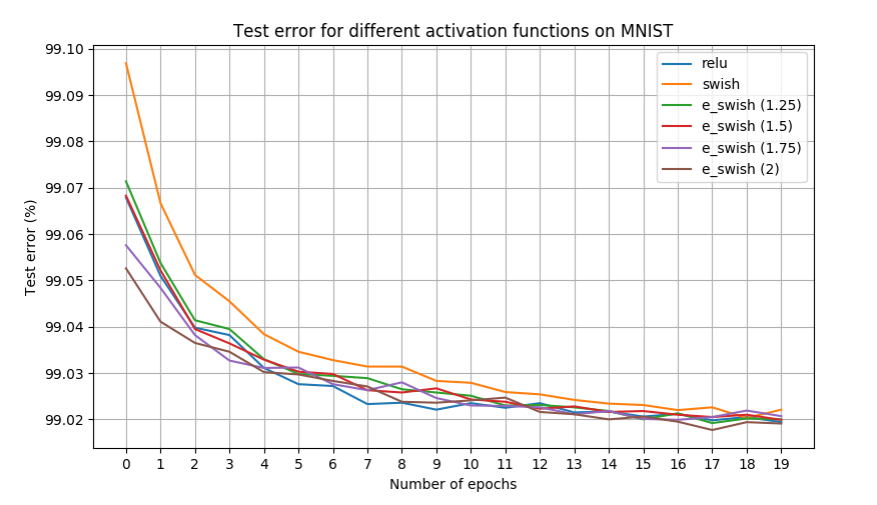}
\caption{Test accuracy (median of 3 runs) on MNIST for fully connected networks with different activations which depths go from 23 to 44 layers. Best viewed in color.}
\label{fig:mnist_median_3}
\end{figure}
\noindent
\\
As it can be seen in Figure 7, E-swish provides a faster learning and a better overall accuracy than Swish. We can also see that higher values of \(\beta\) provide a faster learning.

\subsubsection{CIFAR}
The CIFAR datasets (Krizhevsky \& Hinton, 2009\cite{krizhevsky_et_al}) consist of 60k colored small images with 32*32 pixels each one. Both CIFAR datasets (Cifar10 and Cifar100) are composed of 50k images for training and 10k for testing purposes. The Cifar10 dataset contains images of 10 different classes such as dog, cat, boat and plane. The Cifar100 datset contains images of 100 different classes, but it requires much fine-grained recognition when compared to Cifar10 given that some classes are very similar.
\\
For Cifar10, we measure the performance of E-swish relative to other activations on 2 different models.
\\
\begin{itemize}
  \item WIDE RESIDUAL NETWORK 10-2
\end{itemize}
First, we compare the different activations on a simple CNN model (Convolutional Neural Network) to show the difference on the performance for the different activations. The WRN 10-2 model achieves 89\% accuracy on Cifar10 with the Relu activation function and the original paper implementation(Zagoruyko \& Komodakis, 2016\cite{zagoruyko_and_komodakis}) while having only 310k parameters.
\\
We use the same optimizer as the original implementation (SGD) but change the learning rates in order to speed up training from 200 epochs to 125. The initial learning rate is set to 0.125 and we multiply it by 0.2 at 45, 85 and 105 epochs. We expect that this learning rate setting will provide a faster learning, but it can also have a little bad effect in performance, especially for Swish and E-swish.

\begin{figure}[h!]
\centering
\includegraphics[scale=0.35]{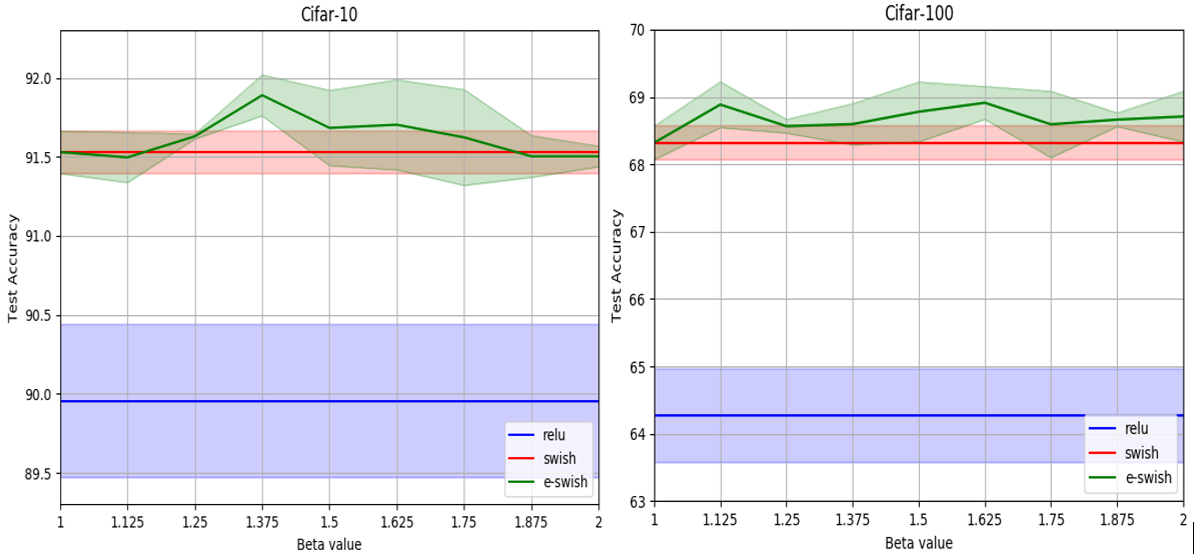}
\caption{Value of parameter β for E-swish with respect to Relu and Swish with 99\% confidence intervals (average of 3 runs). Tests run on CIFAR-10 and CIFAR-100 datasets using WRN-10-2 and standard data augmentation. Color blue represents Relu, red represents Swish and green represents E-swish. Best viewed in color}
\label{fig:cifar_wrn10_2}
\end{figure}
\noindent
\\
As seen in Figure 8 above, E-swish outperforms both Relu and Swish in the range [1.25,1.75] for the \(\beta\) parameter.
\\
\begin{itemize}
  \item DEEPER CNN: SIMPLENET
\end{itemize}
We then compare E-swish against Relu and Swish on a SimpleNet model (Hasanpour et al., 2016\cite{hasanpour_et_al}), which is a deeper CNN composed of 13 convolutional layers. This CNN is designed to achieve a good trade-off between the number of parameters and accuracy, achieving 95\% of accuracy while having less than 6M parameters.
\\
\\
We implemented the model in Tensorflow and replaced the Relu activation function by either Swish or E-Swish. We used the same data augmentation and batch size (128) proposed in the original paper, and used the SGD instead of the Adadelta optimizer used in the original implementation. We ran the model for 150 epochs,with initial learning rate of 0.1 and multiplied it by 0.2 every at 120 epochs.
\\
\\
Since we speculate that values larger than 1.25 for β may cause gradient exploding problems, we only train models with E-swish for values 1.125 and 1.25 for the β parameter.
Due to computational constraints, we were only able to benchmark E-swish against Relu and Swish on single model performance. The results we obtained will be provided below:
\\
\begin{table}[h!]
\centering
\small
\begin{tabular}{lccccccc}
\toprule[0.25mm]
\bf{Activation Functions} & \bf{Test Accuracy(\%)} \\
\midrule
Relu & 95.33 \\
Swish/E-swish (\(\beta = 1\)) & 95.76\\
E-swish (\(\beta = 1.125\))  & \textbf{96.02}\\
E-swish (\(\beta = 1.25\))  & 95.73\\
\bottomrule[0.25mm]
\end{tabular}
\caption{Test accuracy rates for Cifar10 after 150 epochs of training.}
\label{table:simplenet_table}
\end{table}
\\
As Table 3 shows, E-swish outperforms both Relu and Swish. The improvement provided by E-swish is 0.7\% relative to Relu and 0.3\% relative to Swish. Considering that hyperparameters were designed for Relu and that E-swish/Swish were introduced as a drop-in replacement, it is really significative.
\\
\begin{itemize}
  \item WIDE RESIDUAL NETWORK 16-4
\end{itemize}
Finally, we compare E-swish against Relu on a Wide Residual Network 16-4 model (Zagoruyko \& Komodakis, 2016\cite{zagoruyko_and_komodakis}). We have chosen the WRN 16-4 with no dropout since we were not able to train bigger models due to computational constraints. We implemented the model in the Tensorflow framework and replaced the original Relu by the E-swish activation function. We used the same optimizer as the original implementation (SGD) and trained the models for 200 epochs with the same learning rate schedule plus another drop in the learning rate at 180 epochs.

\noindent
\textbf{\underline{Cifar10:}}
\\
Our experiments showed that replacing Relu by Swish or E-swish provided none or minimal benefit on Cifar10. However, this is not the case when the activations are benchmarked on the Cifa100 dataset.
\\ 
\\
\textbf{\underline{Cifar100:}}
\\
Both Swish/E-swish presented an improvement in performance when compared to Relu for the WRN 16-4 model on a single model performance.
\\
\begin{table}[h!]
\centering
\small
\begin{tabular}{lccccccc}
\toprule[0.25mm]
\bf{Activation Functions} & \bf{Test Accuracy(\%)} \\
\midrule
Relu & 75.92\\
Swish/E-swish (\(\beta = 1\)) & 76.88\\
E-swish (\(\beta = 1.25\))  & \textbf{77.35}\\
E-swish (\(\beta = 1.5\))  & 77.25\\
E-swish (\(\beta = 1.75\))  & 77.29\\
E-swish (\(\beta = 2\))  & 77.14\\
\bottomrule[0.25mm]
\end{tabular}
\caption{Test accuracy rates for Cifar100 after 200 epochs of training.}
\label{table:wrn_table}
\end{table}
\\
As shown in Table 2, E-swish provided an improvement of 1.4\% relative to Relu and 0.5\% when compared to Swish. We can also observe that larger values for the β parameter diminished the improvements provided by E-swish.
\\

\section{Discussion and Further Work}
\noindent
As shown in our experiments, values of the \(\beta\) parameter that often provide improvements over both Relu and Swish are in the range of \(1 \leq \beta \leq 2\). We speculate this is due to gradient exploding when \(\beta \leq 2\) and gradient vanishing problems when \(\beta \leq 1\).
\\
\\
In this work, we’ve referred to β as a paramenter but, in fact, it can be considered as a hyperparameter, since it is non-learnable and the choice of β directly affects performance. Empirically, we found that values for β around 1.5 worked well for shallow and mid-scale networks. For large CNNs, we found that 1.125 and 1.25 worked better (SimpleNet). However, these findings are affected by the presence of residual connections, which lower the difference between the different values of β (WRN 16-4).
\\
\\
Further experiments could not be performed due to the resource-intensive nature of training very deep networks and computational constraints by the time of the publication of this paper. For that reason, we would like to invite everyone to perform their own experiments with E-swish in order to get a more extensive proof of its superiority against other activations. 
Especially, we are interested in further testing of E-swish on Wide Residual Networks, DenseNets (Huang et al., 2016\cite{huang_et_al}), ResNets (He et al., 2015\cite{he_residual}), GANs (Goodfellow et al., 2014\cite{goodfellow_et_al}), Autoencoders, RNNs, NASNets (Zoph et al., 2017\cite{zoph_et_al}), and any other model that achieves state of the art results in order to see if it achieves better results in these models. 
We are also interested in the possible application of E-swish to SNNs (Liu et al., 2017\cite{liu_et_al}).

\section{Conclusion}
In this paper, we have presented a new novel activation function, which we call E-swish, which can be formally described as \(f\left(x\right)=\beta x\cdot sigmoid(x)\).
\\
We also showed that the \(\beta\) parameter depended on the network depth, proving that choosing either a large or a small \(\beta\) harmed performance.
\\
\\
Our experiments used models with hyperparameters that were designed for Relu and just replaced Relu by Swish, E-swish or another activation function. Although this was a suboptimal technique, E-swish outperformed both Swish and Relu on a variety of problems. 
\\
Although we have been unable to achieve State-of-the-Art results due to the resource-intensive nature of training huge models, we have shown that E-swish consistently outperformed other activation functions on a range of different problems, depths and architectures and we have managed to improve an existing result for a relatively big and recent model by changing the original activation (Relu) by E-swish.
\\
\\
Further work could apply this new and novel function to more complex models and produce new State-of-the-Art results for different datasets.
\\
\\
\subsubsection*{Acknowledgements}
We would like to thank Oriol Vinyals for the interest in this paper and the Keras and Tensorflow development teams.

\small
\printbibliography

\end{document}